# Indoor Localization Using Visible Light Via Fusion of Multiple Classifiers

Xiansheng Guo, *Member, IEEE*
Sihua Shao, *Student Member, IEEE*
Nirwan Ansari, *Fellow, IEEE*
Abdallah Khreishah, *Member, IEEE*

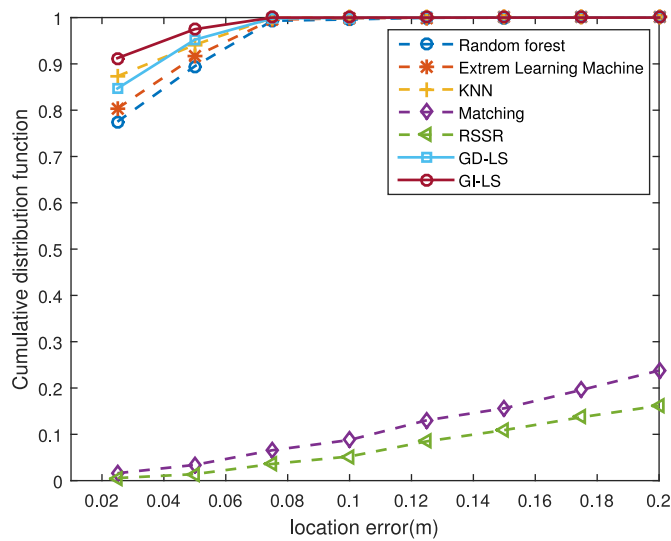

# Indoor Localization Using Visible Light Via Fusion of Multiple Classifiers

**Xiansheng Guo**,[1] *Member, IEEE*,
**Sihua Shao**,[2] *Student Member, IEEE*,
**Nirwan Ansari**,[2] *Fellow, IEEE*,
and **Abdallah Khreishah**,[2] *Member, IEEE*

[1]Department of Electronic Engineering, University of Electronic Science and Technology of China, Chengdu 611731, China
[2]Department of Electrical and Computer Engineering, New Jersey Institute of Technology, Newark, NJ 07102 USA

DOI:10.1109/JPHOT.2017.2767576
1943-0655 © 2017 IEEE. Translations and content mining are permitted for academic research only.
Personal use is also permitted, but republication/redistribution requires IEEE permission.
See http://www.ieee.org/publications_standards/publications/rights/index.html for more information.

Manuscript received June 25, 2017; revised October 24, 2017; accepted October 25, 2017. Date of publication October 30, 2017; date of current version November 14, 2017. This work was supported in part by the National Natural Science Foundation of China under Grants 61371184, 61671137, 61771114, and 61771316, and in part by the Fundamental Research Funds for the Central Universities under Grant ZYGX2016J028. (Corresponding author: Xiansheng Guo).

**Abstract:** We propose a localization technique by fusing multiple classifiers based on received signal strengths (RSSs) of visible light in which different intensity-modulated sinusoidal signals emitted by LEDs are captured by photodiodes placed at various grid points. First, we obtain some approximate RSSs fingerprints by capturing the peaks of power spectral density of the received signals at each given grid point. Unlike the existing RSSs-based algorithms, several representative machine learning algorithms are adopted to train multiple classifiers based on these RSSs fingerprints. Then, two robust fusion localization algorithms, namely, grid-independent least square and grid-dependent least square (GD-LS), are proposed to combine the outputs of these classifiers. A singular value decomposition (SVD)-based LS (LS-SVD) method is proposed to mitigate the numerical stability problem when the prediction matrix is singular. Experiments conducted on an intensity-modulated direct detection system show that the probability of having mean square positioning error of less than 5 cm achieved by GD-LS is improved by 93.03% and 93.15%, respectively, as compared to those by the RSS ratio and RSS matching methods with the fast Fourier transform length of 2000.

**Index Terms:** Indoor positioning, visible light communications (VLC), received signal strengths (RSSs) fingerprints, intensity modulated direct detection (IM/DD), machine learning, fusion localization.

## 1. Introduction

Indoor localization of mobile devices is critical in facilitating myriad location based applications including indoor navigation as well as location-aware services and advertisements in large public buildings such as museums or shopping malls. Although the global positioning system (GPS) has been widely used for precise outdoor localization, it is often unavailable in indoor environments where signals from satellites are strongly attenuated or affected by multipath propagation.

There exists a number of techniques to provision indoor localization; typical examples include systems based on WiFi, ultrawideband, radio frequency identification (RFID), Bluetooth, and ultrasound [1]–[4]. These technologies have limitations such as requiring additional infrastructure, low accuracy, electromagnetic interference, low security, and long response. Comparatively speaking, white light emitting diodes (LEDs) have been widely used because of their durability, eco-friendliness, and lighting efficiency [5]. Visible light communications (VLC) based on white LEDs can be utilized for both lighting and communications simultaneously [6]–[8]. Additionally, the technique utilizes the unlicensed free spectrum and does not incur electromagnetic interference, and thus VLC based on LEDs has been studied in various fields such as illumination, broadcasting, and intelligent transportation systems [9]. Therefore, indoor positioning systems using VLC have recently gained popularity as effective alternatives to the traditional techniques [10].

In this paper, a novel multiple classifiers fusion localization system is proposed based on some RSSs fingerprints. We first transmit different sinusoidal signals that are intensity modulated by four LEDs with known locations, and the signals are received by a Photo Diode (PD). We divide a location area into $G = q \times q$ grid points. Then, we build the RSSs fingerprints by capturing the peaks of power spectral density (PSD) of the received signals at each grid point. Several representative machine learning approaches are studied to train multiple classifiers based on the RSSs fingerprints. We show that the multiple classifiers estimators outperform the classical RSS-based LED localization approaches in accuracy and robustness. Consider that each classifier is accurate for some RSSs pattern, and no particular classifier is universally better than all the others. To further improve the localization performance, two robust fusion localization algorithms, namely, grid independent least square (GI-LS) and grid dependent least square (GD-LS), are proposed to combine the outputs of these classifiers. Considering that the LS based fusion will suffer from numerical stability when the prediction matrix is singular, we also propose a singular value decomposition (SVD) based LS (LS-SVD) solution to mitigate this problem. Experiments conducted on intensity modulated direct detection (IM/DD) system demonstrate the effectiveness of the proposed algorithms. The results show that the proposed algorithms outperform some existing RSS-based algorithms as well as any single classifier based localization algorithm.

The main contributions of this work are summarized as follows:

1) We first propose a machine learning-based LED localization framework. As compared with the conventional RSSs matching and the RSS ratio (RSSR) methods, the multiple classifiers are immune from the high correlation of RSSs fingerprints. The results show that the probability of having mean square positioning errors (MSPEs) of the multiple classifiers of less than 5 cm is improved by 79.76% and 79.88%, respectively, as compared to RSSR and RSS matching methods, with the FFT length being 2000.
2) Two novel fusion methods, namely, grid dependent least square (GD-LS) and grid independent least square (GI-LS), are proposed to obtain a more accurate localization result by taking advantages of each classifier. The proposed algorithms show good numerical stability which is attributed to the use of the LS-SVD solution.
3) The proposed system adopts a relative localization framework and does not need to know the exact locations of the four LEDs. Although the peaks of PSD cannot yield the exact power estimates, our proposed localization framework is more robust to the RSSs estimate biases.
4) The proposed LED localization system does not need to estimate many parameters of the channel model. So, it is more robust to model errors.

The rest of the paper is organized as follows. Section 2 presents some related studies about VLC localization. The proposed system is discussed in Section 3. We detail our proposed algorithms in Section 4. The experimental setup and results are illustrated in Section 5 and some conclusions are drawn in Section 6.

## 2. Related Works

Owing to the many advantages of LEDs, a variety of techniques for indoor positioning using LEDs, have been proposed, including received signal strengths (RSSs) [11]–[13], angle-of-arrival (AOA) [14], time-of-arrival (TOA) [15], time-difference-of-arrival (TDOA) [16], image [17], [18], and their combinations [19]. Among them, AOA achieves a very good accuracy. The main disadvantage is the use of a sensor array, which is expensive. Because of the very short traveling time of the signals for indoor environments due to the short distances, TOA will require a precise synchronization between the receiver and the LEDs. TDOA requires synchronization between the LEDs and thus incurs an increased cost for the installation of the positioning system. Image-based positioning techniques suffer from low accuracy of localization because of some errors induced by image processing algorithms.

RSS-based approaches do not need the synchronization between the LEDs. Most of the existing RSS-based LED localization approaches need to estimate the distances between transmitters and a PD by using the received RSSs. It is well known that the received RSSs are affected by many model parameters, such as radiation and incidence angles, model order, model type, detector physical area, gain of the optical filter and the distance between transmitters and PD [11]. So, it is difficult to estimate the distance accurately between transmitters and a PD, which will degrade the localization performance of some trilateration based localization techniques. Jung *et al.* [20] proposed an RSS ratio (RSSR) location technique, which uses strength ratio between received signals to obtain the distance ratio and the final location estimate can be obtained by solving the distance ratio-based equations. The RSSR method needs to know the model order and several assumptions about the channel need to be made to simplify the derivation of the algorithm. Hann *et al.* [21] proposed a correlation sum ratio (CSR) location technique which uses the correlation between the received data and the assigned addresses to the different LEDs. The extinction ratio (ER) localization algorithm is based on the On Off Key (OOK) modulated signals [22], [23]. The methods in [21]–[23] need some special modulation information to differentiate from which LEDs these signals are emitted. Additionally, it is well known that the performance will degenerate seriously when the model assumption does not hold [11], [12].

Machine learning has been extensively studied in the radio frequency (RF)-based indoor localization, such as WiFi, ZigBee, and Ultra Wideband [24]–[27]. Machine learning based methods have been shown to outperform the traditional RSS-based approaches in accuracy and robustness in coping with the fluctuations of RSSs. The information fusion based positioning strategy has shown good performance in Bluetooth, GSM and WLAN environments [28], [29]. The weighing strategy in [28] severely depends on the correlation between the testing sample and the training samples, and is thus not suitable for the fingerprints with higher correlations. The method in [29] will suffer from numerical instability when its prediction matrix is singular. However, to our best knowledge, machine learning based fusion has not been exploited to facilitate LED based localization.

In this paper, we first propose a machine learning based indoor LED localization framework based on some inaccurate RSSs fingerprints. As compared with some existing RSS-based LED localization approaches, the machine learning based localization approach shows high accuracy and is more robust to model errors, inaccurateness, and high correlations of RSSs. To further improve the localization performance, two robust fusion localization algorithms, namely, grid independent least square (GI-LS) and grid dependent least square (GD-LS), are proposed to combine the outputs of these classifiers. We also use a singular value decomposition (SVD) based LS (LS-SVD) method to mitigate the numerical stability problem when the rank of $\hat{X}$ (Eq. (10)) is smaller than $\mathcal{H}$, which will be discussed in Section 4). The fusion strategies can exploit the complementary information of multiple classifiers to enhance performance in local regions and at the same time reduce the risk of selecting a poor classifier.

## 3. Proposed System

### 3.1 Signal Model

In this section, we propose a simple VLC positioning platform based on RSSs, which are obtained from the peaks of power spectral density (PSD) of received signals. The system configuration is shown in Fig. 1. Assume that we can transmit $M$ different sinusoidal signals $s_i(t)$ with different frequencies $f_i$ from $M$ LEDs transmitters at positions $\mathbf{z}_i = [a_i, b_i, h]^T$, as illustrated in Fig. 2. The signals received by the receiver at a location with unknown position $\mathbf{p} = [x, y, 0]^T$ can be expressed

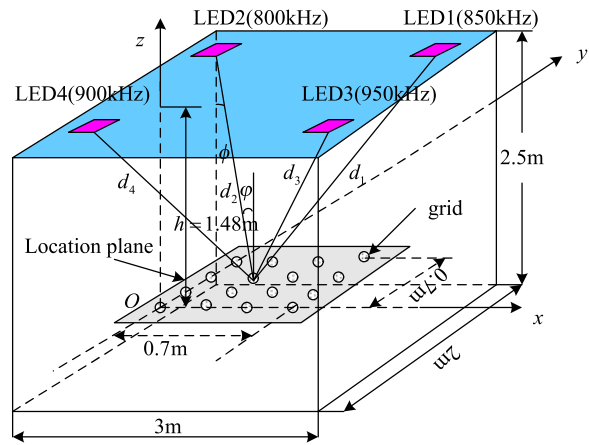

Fig. 1. System configuration.

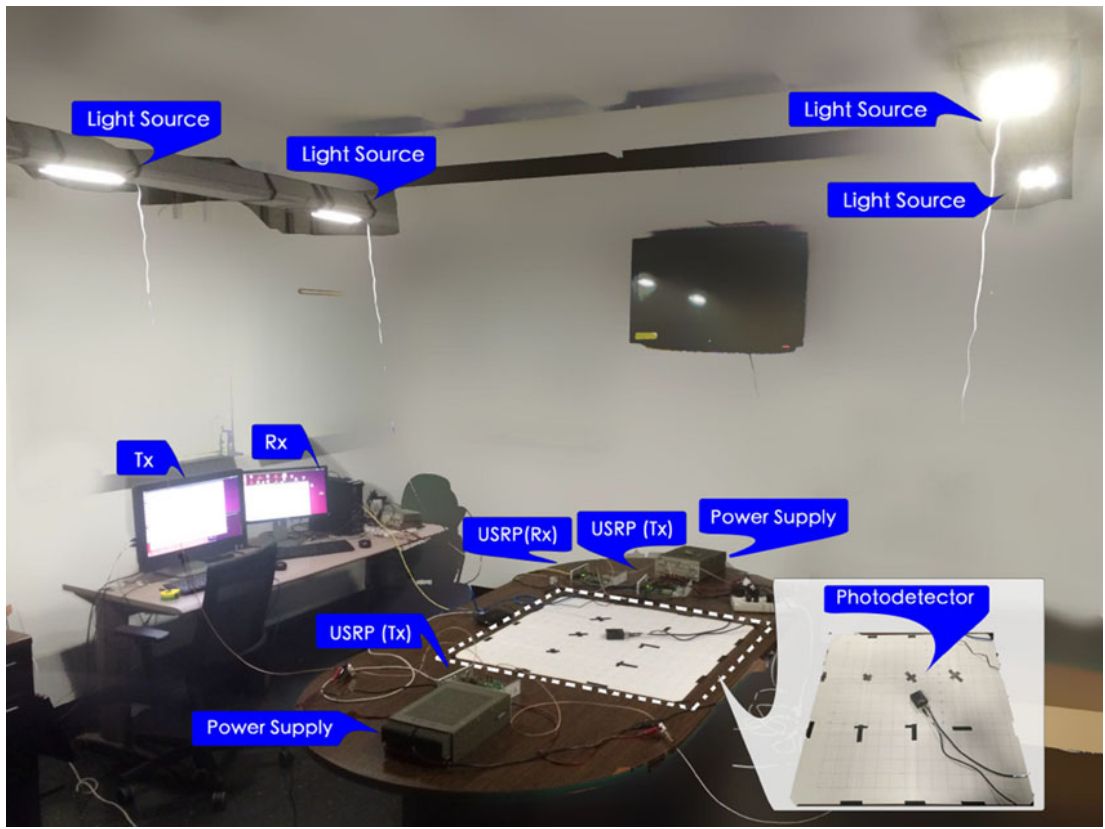

Fig. 2. The experimental testbed.

as

$$y(t) = \sum_{i=1}^{M} \alpha_i \beta_i s_i(t - \tau_i) + n(t), \quad (1)$$

where $\alpha_i$ is the signal attenuation of the optical channel between the *i*th LED and the PD in the scenario as shown in Fig. 1. The responsivity $\beta_i$ is the conversion factor from the optical to the electrical domain. It is a function of the wavelength of the light received. $n(t)$ is the noise. $\alpha_i$ and $\beta_i$ can be treated as constant over the optical bandwidth of the ranging signal. For the line-of-sight (LOS) environment, given a generalized Lambertian LED with order $m$, $\alpha_i$ is a function of the area of the photodiode $S$, the distance $d_i$ between each LED and receiver, and the radiation and incidence angles $\phi$ and $\varphi$ with respect to the transmitter and receiver, as shown in (2)

$$\alpha_i = \frac{(m+1)S}{2\pi d_i^2} \cos^m \phi \cos \varphi. \tag{2}$$

The time delay $\tau_i = d_i/c$, where $c$ is the speed of light and the distance $d_i$ between the $i$th LEDs and the receiver is

$$d_i = \sqrt{\left(h^2 + (x_i - x)^2 + (y_i - y)^2\right)}. \tag{3}$$

Now, we consider the case where $s_i(t)$ is a DC-biased windowed sinusoid waveform with duration $T$ and is given by

$$s_i(t) = w(t) u_i(t) = w(t) + w(t) \cos(2\pi f_i t), \tag{4}$$

where $u_i(t) = 1 + \cos(2\pi f_i t)$. The first term of (4) is a baseband component and the second term is a bandpass component centered at $f_i$.

The periodogram power spectral density (PSD) estimate of $y(t)$ can be expressed as

$$\hat{S}_{\text{per}}(\omega) = \frac{1}{N} |Y_N(\omega)|^2, \tag{5}$$

where $N$ is the length of FFT and $Y_N(\omega)$ is the FFT of $y(t)$, which can be expressed as

$$Y_N(\omega) = \sum_{t=0}^{N-1} y(t) e^{-j\omega t}. \tag{6}$$

It is well known that peaks of $\hat{S}_{\text{per}}(\omega)$ indicate the average powers of the received signals at different frequencies. As a result, we can obtain $M$ RSS values of each received signal by capturing the peaks of fixed frequency locations in the estimated PSD. That is, the RSS vector can be given by

$$\mathbf{r} = \left[\hat{S}_{\text{per}}(f_1), \hat{S}_{\text{per}}(f_2), \ldots, \hat{S}_{\text{per}}(f_M)\right]^T, \tag{7}$$

where $[\cdot]^T$ is the transpose operator.

### 3.2 RSS Fingerprints Construction

Assume we can represent a location area by a grid of $G = q \times q$. At each grid point, a PD receives the signals transmitted from $M$ LEDs and combines them into $y(t)$. Based on the received $y(t)$, $(t = 1, 2, \ldots, \mathcal{T}, \mathcal{T} \gg N)$, we can compute $\mathcal{Q} = \mathcal{T}/N$ PSDs at each point by using Eqs. (5) and (6). Then, the RSS vectors at each point can be obtained by using (7). This process is called site-survey. We can repeat this process $G$ times to build all the fingerprints. Denote the final RSS fingerprints as $\mathbf{R}$; we can summarize the procedures of constructing RSS fingerprints in Algorithm 1.

## 4. Proposed Algorithm
### 4.1 Overview of the Proposed Algorithm

This paper proposes two least square (LS) based multiple classifiers fusion algorithms to improve the accuracy of LED localization by using the RSS fingerprints which are inherently inaccurate. The proposed method first uses advanced machine learning approaches to estimate the positions of a receiver. Then, two LS based fusion methods are proposed to further improve the accuracy of localization. Unlike traditional methods, the proposed approach, by leveraging each classifier

---

**Algorithm 1: The RSS fingerprints construction.**

**Input:** 1) The number of grid points $G$; 2) $\mathcal{T}$ points of $y(t), t = 1, 2, \cdots, \mathcal{T}$ from each grid point; 3) The length of FFT $N$ at each grid point.

**Output:** The RSS fingerprints $R$.

1: **for** $g = \{1, 2, \cdots, G\}$ **do**
2:   **for** $k = \{1, 2, \cdots, Q\}$ **do**
3:     Compute $N$-length FFT by using (6);
4:     Compute the estimate of PSD using (5);
5:     Compute the RSS vector $r^k$ using (7);
6:   **end for**
7:   $R_g = [r^1, r^2, \cdots, r^Q]$;
8: **end for**
9: $R = [R_1, R_2, \cdots, R_G]$;
10: **return** $R$

---

being tuned to recognize specific input patterns, uses the complementary advantages of multiple classifiers, weighs the various estimation results, and combines them to improve accuracy.

Assume that we have $\mathcal{H}$ classifiers available. Given the input RSSs fingerprinting vector $r_j$ collected from the real location $[x_g, y_g, 0]^T$ at the $g$th grid point. Denote the location estimate as

$$\hat{p}_j^\eta = \left[\hat{x}_j^\eta, \hat{y}_j^\eta, 0\right]^T \triangleq h_\eta(r_j, R), \eta = 1, 2, \ldots, \mathcal{H}, \tag{8}$$

where $j = 1, 2, \ldots, L$ with $L$ being the total mumber of inputted RSSs fingerprinting vectors. $\triangleq$ means that we need to obtain the location estimate by transforming binary classification into our multivariant classification framework in VLC indoor localization.

The location estimate vector $\hat{x}$ by the weighed multiple classifier can be given by

$$\hat{x} = \hat{X} w_x, \tag{9}$$

where $w_x = [w_{x1}, w_{x2}, \ldots, w_{x\mathcal{H}}]^T$ is the weights vector of the $x$-axis. $\hat{X}$ is the $L \times \mathcal{H}$ prediction matrix:

$$\hat{X} = \begin{bmatrix} \hat{x}_1^1 & \hat{x}_1^2 & \cdots & \hat{x}_1^{\mathcal{H}} \\ \hat{x}_2^1 & \hat{x}_2^2 & \cdots & \hat{x}_2^{\mathcal{H}} \\ \vdots & \vdots & \ddots & \vdots \\ \hat{x}_L^1 & \hat{x}_L^2 & \cdots & \hat{x}_L^{\mathcal{H}} \end{bmatrix}, \tag{10}$$

where each row represents the abscissa estimates from the $\mathcal{H}$ classifiers. As a result, the problem is to estimate $w_x$ by utilizing (10) with known $x$ and $\hat{X}$.

Assume that we know the real locations $x = [x_1, x_2, \ldots, x_L]^T$ which corresponds to the inputs vector $r_1, r_2, \ldots, r_L$. The estimated error vector can be expressed as

$$e = |x - \hat{x}| = |x - \hat{X} w_x|. \tag{11}$$

The square error can be written as

$$\|e\| = e^T e = \left(x - \hat{X} w_x\right)^T \left(x - \hat{X} w_x\right). \tag{12}$$

If $L > \mathcal{H}$ and $\hat{X}$ is a nonsingular matrix, by finding the least squares (LS) error, the weights $w_x$ can be given by the following LS solution

$$\hat{w}_x = \left(\hat{X}^T \hat{X}\right)^{-1} \hat{X}^T x, \tag{13}$$

where $(\cdot)^{-1}$ is the matrix inverse. Note that the LS solution $\hat{w}_x$ is not stable when the rank of $\hat{X}$ is smaller than $\mathcal{H}$, which happens when some of the classifiers yield the same prediction for each testing sample. Hence, we can use the singular value decomposition (SVD) to obtain a robust solution.

Assume that the rank of $\hat{X}$, denoted by $K$, is smaller than $\mathcal{H}$, then the SVD of $\hat{X}$ can be expressed as

$$\hat{X} = U \begin{bmatrix} \Sigma_K & \mathbf{0} \\ \mathbf{0} & \mathbf{0} \end{bmatrix} V^T, \tag{14}$$

where $\Sigma_K = \text{diag}(\sigma_1, \sigma_2, \ldots, \sigma_K)$ is the singular values matrix with $\sigma_\kappa$ being the $\kappa$th singular value. $U = [u_1, u_2, \ldots, u_L]$ and $V = [v_1, v_2, \ldots, v_\mathcal{H}]$ are the corresponding left and right singular vectors, respectively. Also, we can rewrite (14) based on the rank of $\hat{X}$ as

$$\begin{aligned}\hat{X} &= U \begin{bmatrix} \Sigma_K & \mathbf{0} \\ \mathbf{0} & \mathbf{0} \end{bmatrix} V^T \\ &= [U_1 \; U_2] \begin{bmatrix} \Sigma_K & \mathbf{0} \\ \mathbf{0} & \mathbf{0} \end{bmatrix} \begin{bmatrix} V_1^T \\ V_2^T \end{bmatrix} \\ &= U_1 \Sigma_K V_1^T, \end{aligned} \tag{15}$$

in which $U_1$, $U_2$, $V_1$ and $V_2$ are the left and right submatrice corresponding to the signals and noise, respectively. By substituting (15) into (13), we can obtain the LS-SVD based weights estimate as

$$\hat{w}_x = U_1 \Sigma_K^{-1} V_1^T x. \tag{16}$$

Consider that the solution of (16) contains both the left and the right singular vectors, we can further simplify (16) by only using the left singular vectors $v_k$. Note that $(\hat{X}^T \hat{X})^{-1} = \sum_{\kappa=1}^{K} \frac{v_\kappa v_\kappa^T}{\sigma_\kappa^2}$; substituting this into (13), we obtain

$$\hat{w}_x = \sum_{\kappa=1}^{K} \frac{v_\kappa v_\kappa^T}{\sigma_\kappa^2} \hat{X}^T x = \sum_{\kappa=1}^{K} \left( \frac{v_\kappa^T \theta}{\sigma_\kappa^2} \right) v_\kappa, \tag{17}$$

where $\theta$ is the cross correlation vector between $\hat{X}$ and $x$, and can be written as

$$\theta = \hat{X}^T x. \tag{18}$$

Similarly, the weights vector $\hat{w}_y$ can be estimated by repeating the above procedures.

### 4.2 Weights Selection Strategies

Here, we propose two alternative weights selection strategies when given an offline testing RSS fingerprints vector $r_j$: "*grid-independent(GI)*" method and the "*grid-dependent(GD)*" method.

*4.2.1 Grid-Independent (GI) Weights Selection Approach:* The idea of the GI approach for selecting weights is that we only use fixed weights to fuse the final estimated location. That is, in the weights estimation stage, we combine all the $L$ samples from different grid points to build one $\hat{X}$, and all the real locations corresponding to these $L$ samples are collected to form one $x$. The weights $\hat{w}_x$ and $\hat{w}_y$ can be estimated by using (13) or (18) based on the singularity or non-singularity of $\hat{X}$. The final location estimate can be given by (9).

*4.2.2 Grid-Dependent (GD) Weights Selection Approach:* GD tries to assign different weight vectors to different grid points. In this case, we should collect the $L$ testing samples $r_j$ for each point to estimate the corresponding weights by using (9). Based on the singularity or non-singularity of $\hat{X}$, we can obtain two $\mathcal{H} \times G$ weights matrice $W_x$ and $W_y$, respectively, which can be expressed as

$$\hat{W}_x = \left[\hat{w}_x^1, \hat{w}_x^2, \ldots, \hat{w}_x^G\right]^T \quad (19)$$

$$= \begin{bmatrix} \hat{w}_{x1}^1 & \hat{w}_{x1}^2 & \cdots & \hat{w}_{x1}^G \\ \hat{w}_{x2}^1 & \hat{w}_{x2}^2 & \cdots & \hat{w}_{x2}^G \\ \vdots & \vdots & \ddots & \vdots \\ \hat{w}_{x\mathcal{H}}^1 & \hat{w}_{x\mathcal{H}}^2 & \cdots & \hat{w}_{x\mathcal{H}}^G \end{bmatrix}$$

and

$$\hat{W}_y = \left[\hat{w}_y^1, \hat{w}_y^2, \ldots, \hat{w}_y^G\right]^T \quad (20)$$

$$= \begin{bmatrix} \hat{w}_{y1}^1 & \hat{w}_{y1}^2 & \cdots & \hat{w}_{y1}^G \\ \hat{w}_{y2}^1 & \hat{w}_{y2}^2 & \cdots & \hat{w}_{y2}^G \\ \vdots & \vdots & \ddots & \vdots \\ \hat{w}_{y\mathcal{H}}^1 & \hat{w}_{y\mathcal{H}}^2 & \cdots & \hat{w}_{y\mathcal{H}}^G \end{bmatrix},$$

where $\hat{w}_x^g = \left[\hat{w}_{x1}^g, \hat{w}_{x2}^g, \ldots, \hat{w}_{x\mathcal{H}}^g\right]^T$ and $\hat{w}_y^g = \left[\hat{w}_{y1}^g, \hat{w}_{y2}^g, \ldots, \hat{w}_{y\mathcal{H}}^g\right]^T$ are the weights vector of the $g$th grid point.

In the online stage, we need to determine the weights on which grid point will be selected; an available weights selection method is the minimization of Euclidean distance between the input testing sample $\tilde{r}$ and the mean vectors of the fingerprints on each grid point.

$$\hat{g} = \arg\min_{g \in \{1, 2, \ldots, G\}} \|\tilde{r} - R_g\|_2, \quad (21)$$

where $\|\cdot\|_2$ is the $\ell_2$-norm.

The final location estimate by GD is given by

$$\begin{cases} \hat{x} = \hat{X} \hat{w}_x^g \\ \hat{y} = \hat{Y} \hat{w}_y^g \end{cases} \quad (22)$$

Given an online testing RSS vector $\tilde{r}$, now we can summarize the procedures of the GI-LS and GD-LS fusion localization algorithms in Algorithms 2 and 3, respectively.

### 4.3 Classifier Functions

Machine learning has been applied to many interesting problems in many fields, such as image processing, machine vision, and signal processing. Each machine learning approach can work as a classifier function $h_i(R)$ to yield an expertise prediction. We consider three typical classifiers; extreme learning machine (ELM), Random forest, and K-neighbour nearest (KNN).

*4.3.1 ELM:* ELM is based on the empirical risk minimization theory and makes use of a single-layer feedforward network for the training of single hidden layer feedforward neural networks (SLFNs) [24]. The learning process needs only one single iteration and avoids multiple iterations and local minimization. In VLC localization, the goal of ELM is to find the relationship between the RSSs fingerprints and its corresponding location label.

The prediction of a standard SLFNs with $\mathcal{L}$ hidden neurons could be mathematically modeled by:

$$h_{\text{ELM}}(\tilde{r}, R) = \text{sign}\left(\sum_{v=1}^{\mathcal{L}} \tilde{w}_v \tilde{f}_v(\tilde{r}, R)\right), \quad (23)$$

---

**Algorithm 2: GI-LS fusion localization algorithm.**

**Input:** 1) The online testing RSS vector $\tilde{r}$; 2) The $L$ offline testing RSS vector $r_i$; 3) The offline training fingerprints $R$; 4) The $\mathcal{H}$ different classifiers $h_\eta(R)$.
**Output:** The location estimate $[\hat{x}, \hat{y}]^T$.
1: **The offline phase:**
2: **for** $j = \{1, 2, \cdots, L\}$ **do**
3: 　**for** $\eta = \{1, 2, \cdots, \mathcal{H}\}$ **do**
4: 　　Compute $\hat{p}_j^\eta = h_\eta(r_j, R)$ by using (8);
5: 　**end for**
6: **end for**
7: Obtain $\hat{X}$ and $\hat{Y}$ using (10);
8: Compute $\hat{w}_x$ and $\hat{w}_y$ using (17);
9: **The online phase:**
10: **for** $\eta = \{1, 2, \cdots, \mathcal{H}\}$ **do**
11: 　Compute $\hat{p}^\eta = h_\eta(\tilde{r}, R)$;
12: **end for**
13: Obtain $\hat{X}$ and $\hat{Y}$ using (10);
14: Compute $\hat{x}$ and $\hat{y}$ using (9);
15: **return** $[\hat{x}, \hat{y}]^T$.

---

**Algorithm 3: GD-LS fusion localization algorithm.**

**Input:** 1) The online testing RSS vector $\tilde{r}$; 2) The $L$ offline testing RSS vector $r_j$; 3) The offline training fingerprints $R$; 4) The $\mathcal{H}$ different classifiers $h_\eta(R)$.
**Output:** The location estimate $[\hat{x}, \hat{y}]^T$.
1: **The offline phase:**
2: **for** $g = \{1, 2, \cdots, G\}$ **do**
3: 　**for** $j = \{1, 2, \cdots, L\}$ **do**
4: 　　**for** $\eta = \{1, 2, \cdots, \mathcal{H}\}$ **do**
5: 　　　Compute $\hat{p}_j^\eta = h_\eta(r_j, R)$ by using (8);
6: 　　**end for**
7: 　**end for**
8: 　Obtain $\hat{X}$ and $\hat{Y}$ using (10);
9: 　Compute $\hat{w}_x$ and $\hat{w}_y$ using (17);
10: **end for**
11: Obtain $\hat{W}_x$ and $\hat{W}_y$ using (19) and (20);
12: **The online phase:**
13: **for** $\eta = \{1, 2, \cdots, \mathcal{H}\}$ **do**
14: 　Compute $\hat{p}^\eta = h_\eta(\tilde{r}, R)$;
15: **end for**
16: Obtain $\hat{X}$ and $\hat{Y}$ using (10);
17: Compute $\hat{g}$ using (21);
18: Find the weights $\hat{w}_x$ and $\hat{w}_y$ based on $\hat{g}$;
19: Compute the location estimate $[\hat{x}, \hat{y}]^T$ using (22);
20: **return** $[\hat{x}, \hat{y}]^T$.

---

where $\tilde{w}_\nu$ is the $\nu$th output weight between the $\nu$th neuron and the output. $\tilde{f}_\nu(\tilde{r}, R)$ is the $\nu$th output of the hidden layer with respect to the input $\tilde{r}$. sign$(\cdot)$ is a sign function. By minimizing the training error as well as the norm of the output weights, we can estimate the weight $\tilde{w}_\nu$ of ELM [24]. Then, the predictions of the ELM classifier can be obtained based on the estimated weights.

*4.3.2 Random Forest:* Random forest (RF) is a combination of tree predictors such that each tree depends on the values of a random vector sampled independently and with the same distribution for all trees in the forest [25]. A tree is a collection of nodes and edges organized in a hierarchical structure. A decision tree can be interpreted as a technique for splitting complex problems into a hierarchy of simpler ones. Forest training is performed by optimizing the parameters of the weak learner at the $\xi$ split node via

$$\hat{\boldsymbol{\theta}}_\xi = \arg\max \mathcal{I}'_\xi, \tag{24}$$

where $\mathcal{I}'_\xi$ is the information gain of the $\xi$th node. $\hat{\theta}_\xi$ denotes the parameters of the weak learner at the $\xi$th split node. Assume $T$ is a trained tree with weak learner parameters $\hat{\theta}_\xi$, tree depth $t_d$, and node number $t_m$; then, the prediction of the tree can be given by

$$h_{\text{RF}}(\tilde{r}, R) = T(\tilde{r}, R). \tag{25}$$

*4.3.3 KNN:* KNN is a type of instance-based learning, or lazy learning, where the function is only approximated locally and all computation is deferred until classification [26]. It is among the simplest of all machine learning algorithms. KNN calculates the distance between $\tilde{r}$ and each vector in the trained data $R$ by using a given distance metric, such as Euclidean distance and Manhattan distance, etc, to determine the location $p$. Then, all the distances are sorted in ascending order and the number of classifications in the first $k$ samples, $C$, is recorded. The number of samples in each classification is $N_c$ ($c = 1, 2, \ldots, C$) and the most votes to estimate the location label $g$ is chosen:

$$\hat{g} = \arg \max_{c=1,2,\ldots,C} N_c. \tag{26}$$

Then, the prediction of the KNN classifier for giving $\tilde{r}$ becomes

$$h_{\text{KNN}}(\tilde{r}, R) = [x_{\hat{g}}, y_{\hat{g}}, 0]^T. \tag{27}$$

### 4.4 Performance Analysis

*4.4.1 Accuracy:* It is well known that the accuracy of the fingerprint-based localization approach is constrained by the distance between adjacent grids. Our proposed fusion localization framework can mitigate this limitation. Two factors can guarantee the high accuracy of the proposed framework. First, the machine learning approach can improve the accuracy of the localization system by being immune to the fluctuation and correlation of RSS fingerprints. Second, the weighed fusion strategy can yield the location estimates that are not on the grid points, thus improving the accuracy without any grid refinement technique.

*4.4.2 Robustness:* The robustness of our proposed localization framework is derived from the following aspects: first, unlike other trilateration methods, the machine learning classifier does not need to estimate model parameters, and possesses good robustness against model error and RSS error, fluctuation, and correlation (We will show that the correlations among real RSS fingerprints are high enough while the traditional trilateration methods incur low accuracy). Second, the proposed fusion strategy has little effect on the performance of a single classifier. It can leverage the complementary advantages of multiple classifiers, weigh the various estimation results, and combine them to improve the robustness. Finally, LS-SVD can guarantee numerical stability when the prediction matrix $\hat{X}$ in (9) is singular, thus further improving the robustness of our proposed approach.

*4.4.3 Complexity:* The main burden of fingerprint-based localization approach are the fingerprints construction, also known as a site survey. The machine learning approach needs to train these fingerprints prior to executing the online localization. However, these tedious tasks are done offline. Meanwhile, a number of pioneering works, such as Transfer Learning (TL) [27], Matrix Completion (MC) [30], and Compressive Sensing (CS) [31], can be used to reduce the burden of a site survey. In other words, the complexity of our proposed approach mainly comes from the online phase, which is mainly dominated by SVD of $\hat{X}$. Consider that the dimension of $\hat{X}$ in (9) is $L \times \mathcal{H}$, and so its complexity is $\mathcal{O}(L\mathcal{H}^2)$. Since $\mathcal{H} \ll L$, we can obtain a fast location estimate if $L$ is not too large.

## 5. Experimental Setup and Results

Our testbed is built in the optimized networking laboratory (ONL), which is located at the fourth floor of Faculty Memorial Hall (FMH) building, New Jersey Institute of Technology (NJIT). The testbed consists of 4 visible light sources ($M = 4$), each light source contains two LED arrays (premium daylight white LED light bar from Solid Apollo LED). We use two USRPs N210 with two LFTX daugtherboards to drive the light sources for signal transmission and one USRP N210 with one LFRX daughterboard to perform signal reception. The optical to electrical conversion is implemented by an avalanche photodector APD 130 A2/M from Thorlabs. No lens is equipped on the photodetector. Four different sinusoidal signals (i.e., with frequencies 800 kHz, 850 kHz, 900 kHz and 950 kHz, respectively) are generated from GNURadio and forwarded to the USRPs for signal transmission. The sinusoidal signals are then combined with 23.6 Volts DC power supply by 4 Bias tees ZFBT-6GW+ and used to drive 4 visible light sources, as shown in Fig. 2. The localization experiment is performed within a 0.7 m × 0.7 m square area. The height of four LEDs $h$ is 1.48 m. The location of the first grid point is taken as the origin. The locations of four LEDs are $\mathbf{z}_1 = [1.56 \text{ m}, 0.7 \text{ m}, 1.48 \text{ m}]^T$, $\mathbf{z}_2 = [-1.13 \text{ m}, 0.67 \text{ m}, 1.48 \text{ m}]^T$, $\mathbf{z}_3 = [1.56 \text{ m}, -0.47 \text{ m}, 1.48 \text{ m}]^T$, $\mathbf{z}_4 = [-1.13 \text{ m}, -0.5 \text{ m}, 1.48 \text{ m}]^T$, respectively. We record the raw time domain signals at $G = 225$ ($q = 15$) different locations in a grid structure with 5 cm between each pair of adjacent locations. Each location is recorded for 5 seconds. The sampling rate is set to 4 MHz. The time domain samples are then analyzed in MATLAB using the Fast Fourier Transform (FFT) process to extract the signal strengths for constructing RSSs fingerprints by using Algorithm 1. The functional block diagram of our proposed localization system is shown in Fig. 3.

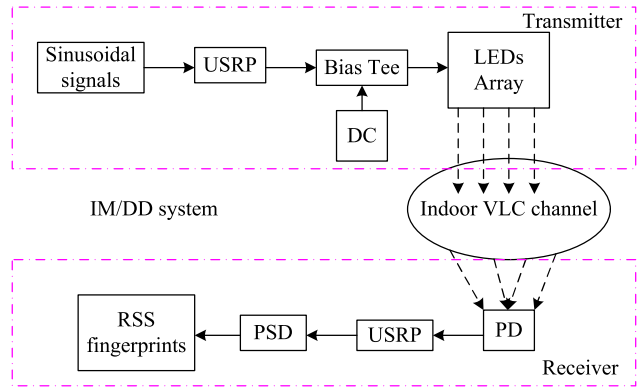

Fig. 3. The functional block diagram of our proposed localization system.

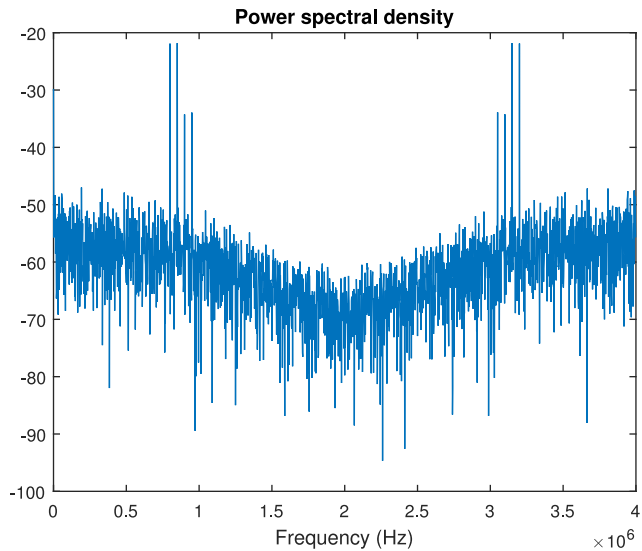

Fig. 4. The power spectral density (PSD) of the received signals of the PD.

TABLE 1
gomezt1-2666154.eps

| FFT length Frequency | 2000 | 4000 | 6000 | 8000 |
|---|---|---|---|---|
| 800kHz | -27.3340 | -24.3208 | -22.5599 | -21.3069 |
| 850kHz | -32.8109 | -29.8088 | -28.0448 | -26.8038 |
| 900kHz | -24.1339 | -21.1263 | -19.3677 | -18.1189 |
| 950kHz | -31.1861 | -28.1668 | -26.3968 | -25.1481 |

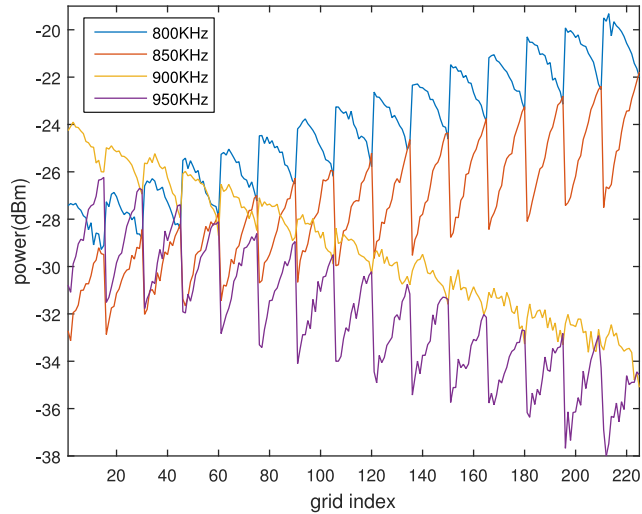

Fig. 5. The power profile of four LEDs at all grid points.

Fig. 4 shows the estimated power spectral density (PSD) of the received signals of the PD when the length of FFT is 2000. From Fig. 4, we find that there are four clear peaks at the corresponding frequencies, which can be captured as the powers of the received signals. It is worth noting that the RSSs from the peaks of the PSD is inaccurate. To show the inaccurateness of the RSSs, we list the mean powers of the four frequencies at the first grid point with FFT lengths of 2000, 4000, 6000, and 8000 in Table 1 (Unit:dBm). The mean RSSs values are computed with 200 samples. From the table, we find that the RSSs are changing with different lengths of FFT even for the same frequency LED. This will degrade the positioning accuracy of the classical RSSs methods, such as RSSR and RSS matching methods. Fortunately, unlike the classical RSSs methods, our proposed algorithms are robust to the inaccurate RSSs.

The length of FFT is set to 2000 ($N = 2000$) Fig. 5 illustrates the instant RSSs of LEDs versus different grid indices. Although the instant RSSs have some slightly fluctuation, the variation trends of RSSs can basically reflect the distances between the PD and the four LEDs. It is seen that the fluctuation of RSSs of LED is smaller than that of the WLAN environment [2]. In other words, the fluctuation is not the main characteristic of LED's RSSs. The main characteristic of the RSSs in VLC is the high correlations among fingerprints, as depicted in Fig. 6, which shows the correlation coefficients between the testing samples and the mean RSS fingerprints at each grid index. It can be seen that no clear peaks appear in the diagonal of the correlation coefficients. It means that there exists higher correlations between each fingerprint vector in the mean RSS fingerprints, which will degenerate the classical RSS matching methods.

To compare the performance of our proposed algorithms with the RSSR method [20] and the classical RSS matching method, the Lambertian model order $m = -\ln 2/\ln \cos(\varphi_{1/2})$ with the semi-angle (at half power) $\varphi_{1/2} = 22°$ is considered. To evaluate the positioning performance of the listed

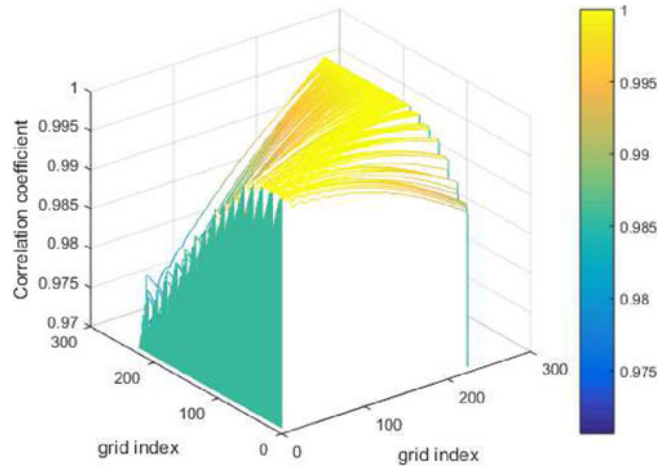

Fig. 6. The correlation coefficients between the testing samples and the mean RSS fingerprints at each grid point.

TABLE 2

The Parameters of the Three Machine Learning Approaches

| | |
|---|---|
| KNN | Distance metric: Euclidean distance; $k = 120$. |
| ELM | Number neurons in input, hidden, and output layers: 4, 600, and 225; Activation function: Sigmoid. |
| Random Forest | Tree depth: 5; Tree number: 40; Weak classifier: Decision stump. |

algorithms, we define the mean square positioning error (MSPE) as

$$\text{MSPE} = \sqrt{\frac{1}{\mathcal{I}} \sum_{i=1}^{\mathcal{I}} \left[ (\hat{x}_i - x)^2 + (\hat{y}_i - y)^2 \right]}, \quad (28)$$

where $\mathcal{I}$ is the number of testing samples and $[\hat{x}_i, \hat{y}_i]$ is the location estimate of the $i$th testing sample. Three machine learning algorithms ($\mathcal{H} = 3$) are compared, including KNN, ELM, and random forest. The parameters implemented in our experiment are listed in Table 2. The positioning results of the three machine learning approaches as well as our proposed fusion algorithms are illustrated in Fig. 7. It is seen that our proposed algorithms and the machine learning approaches perform the LED positioning well.

In order to see their performance clearly, we compare the cumulative distribution functions (CDFs) versus the MSPE for the listed algorithms in Fig. 8. From this figure, we find that the accuracies of the multiple classifiers are superior to the RSSR and the classical matching methods. The probabilities of positioning errors less than 5 cm are 80.4%, 85.56%, and 88.09%, for KNN, ELM, and Random Forest, respectively. While the probabilities of the GD-LS and GI-LS are 88.78% and 93.17%, respectively, implying that the fusion strategy works well in this case. The high correlation between each column in RSS fingerprints, as shown in Fig. 6, has great impact on weights selection of the GD-LS algorithm. In other words, GD-LS will select some wrong weights to fuse because the selected weights are decided by the distance between the testing sample and the training data, as shown in (21). Note that the GD-LS weight selection strategy is similar to the method in [28]. So, it is proven that the GD-LS is not good for VLC fingerprints. As expected, the high correlation directly results in the poor performance of the RSS matching method, as shown in Fig. 8. The probability of the matching based method in acquiring a positioning error of less than 20 cm is only 17.93%. Note that the probability of the GI-LS algorithm in achieving positioning error of less than 7.5 cm has reached 100%, which is better than other compared algorithms. Fig. 9 shows the CDFs versus MSPE when the FFT length $N = 4000$. In comparing Fig. 8 with Fig. 9, we find that our proposed algorithms are robust to the RSSs variation. The longer FFT length, the more robust our proposed algorithms.

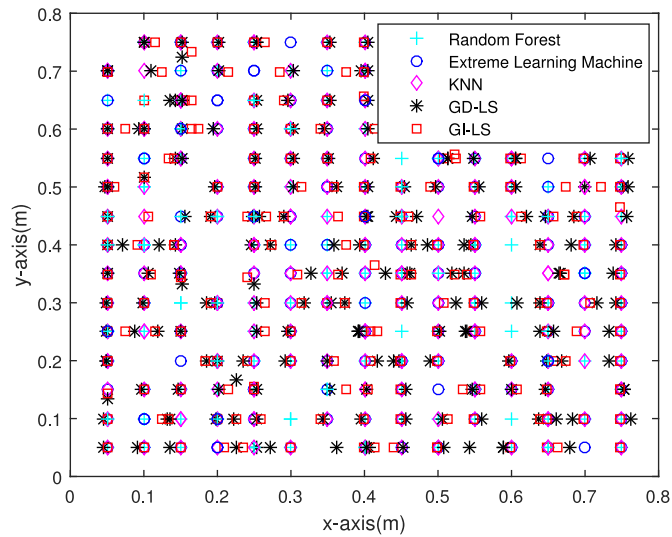

Fig. 7. The positioning results of the compared algorithms in the location plane.

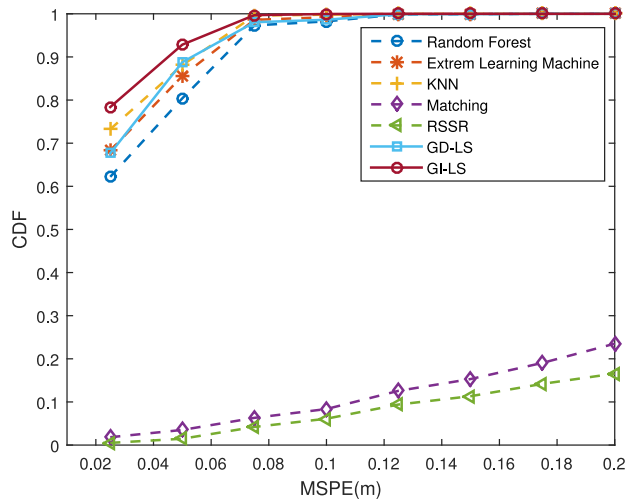

Fig. 8. The CDFs versus MSPE of different localization algorithms, $N = 2000$.

Note that the performance of the RSSR approach is worse than the RSSs matching approach, and probability of its achieving positioning error of less than 20 cm is just 18.65% as shown in Figs. 8 and 9, which is far less than the machine learning approaches and the two fusion algorithms. The high correlation of RSS fingerprints is just one of the reasons that degrades the performance of the RSSR approach. Meanwhile, the RSSR approach needs several assumptions on model parameters. Any parameter mismatch will induce positioning errors. Additionally, the RSSR approach needs to know the locations of the four LEDs accurately. Furthermore, the RSSs captured from the peaks of PSD are not accurate enough, and this is also a bias source of positioning errors. Hence, it is easier to yield wrong location estimates when taking the above reasons into consideration. While the RSSs matching approach is a pattern matching technique, which is slightly robust to the model error but more sensitive to high correlations, it is slightly more robust than the RSSR approach. Comparatively speaking, our proposed method shows high accuracy without knowing model parameters and is robust to high correlations, the biases of LEDs locations, and fluctuations of RSSs.

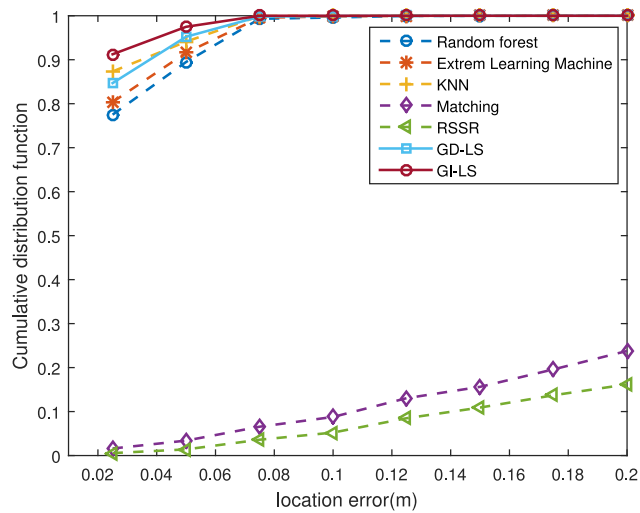

Fig. 9. The CDFs versus MSPE of different localization algorithms, $N = 4000$.

## 6. Conclusion

In this paper, we have proposed a multiple classifiers fusion localization framework by using the RSS fingerprints, which are captured from the peaks of the PSD of received signals. Besides the inaccurateness, the RSSs fingerprints show high correlation, which would introduce some large positioning errors for the classical RSSs matching methods.

To mitigate this problem, we first train the multiple classifiers based on the RSS fingerprints offline. Then, in the online stage, we design two robust fusion algorithms, namely, GD-LS and GI-LS, based on the outputs of these multiple classifiers. The proposed algorithms have been shown to exhibit numerical stability in dealing with singular outputs matrices. Unlike the trilateration techniques based on the RSSs, our proposed algorithms do not depend on the model parameters and are robust to the inaccurateness of the RSSs. Meanwhile, they can also position without knowing the locations of the transmitters. We have validated the feasibility of our proposed positioning framework in the real VLC indoor environment.

Here, we just consider three typical machine learning approaches. Theoretically speaking, the more machine learning algorithms are fused, the more accurate positioning results are obtained. As for the LS based fusion strategy, we can also derive a weighted LS (WLS) solution when the variance of the noise information of the received signals is given [4]. All in all, multiple information fusion is a good strategy to achieve accurate and robust VLC indoor localization.